\documentclass{llncs}
\usepackage{graphicx}
\usepackage{amsmath,amssymb} 
\usepackage{color}
\usepackage[width=122mm,left=12mm,paperwidth=146mm,height=193mm,top=12mm,paperheight=217mm]{geometry}

\usepackage{subfig}
\usepackage{hyperref}
\usepackage{caption}
\usepackage{wrapfig}

\usepackage{multirow}
\usepackage{pifont}
\begin{document}
\pagestyle{headings}
\mainmatter

\title{Piggyback: Adapting a Single Network to Multiple Tasks by Learning to Mask Weights}
\author{Arun Mallya, Dillon Davis, Svetlana Lazebnik}
\institute{University of Illinois at Urbana-Champaign}
\maketitle


\begin{abstract}
This work presents a method for adapting a single, fixed deep neural network to multiple tasks without affecting performance on already learned tasks. 
By building upon ideas from network quantization and pruning, we learn binary masks that ``piggyback'' on an existing network, or are applied to unmodified weights of that network to provide good performance on a new task. 
These masks are learned in an end-to-end differentiable fashion, and incur a low overhead of 1 bit per network parameter, per task. 
Even though the underlying network is fixed, the ability to mask individual weights allows for the learning of a large number of filters. 
We show performance comparable to dedicated fine-tuned networks for a variety of classification tasks, including those with large domain shifts from the initial task (ImageNet), and a variety of network architectures. 
Unlike prior work, we do not suffer from catastrophic forgetting or competition between tasks, and our performance is agnostic to task ordering. \\
Code available at \url{https://github.com/arunmallya/piggyback}.

\end{abstract}


\section{Introduction}
\label{sec:introduction}

\begin{figure*}[t]
\centering
\includegraphics[width=0.9\textwidth]{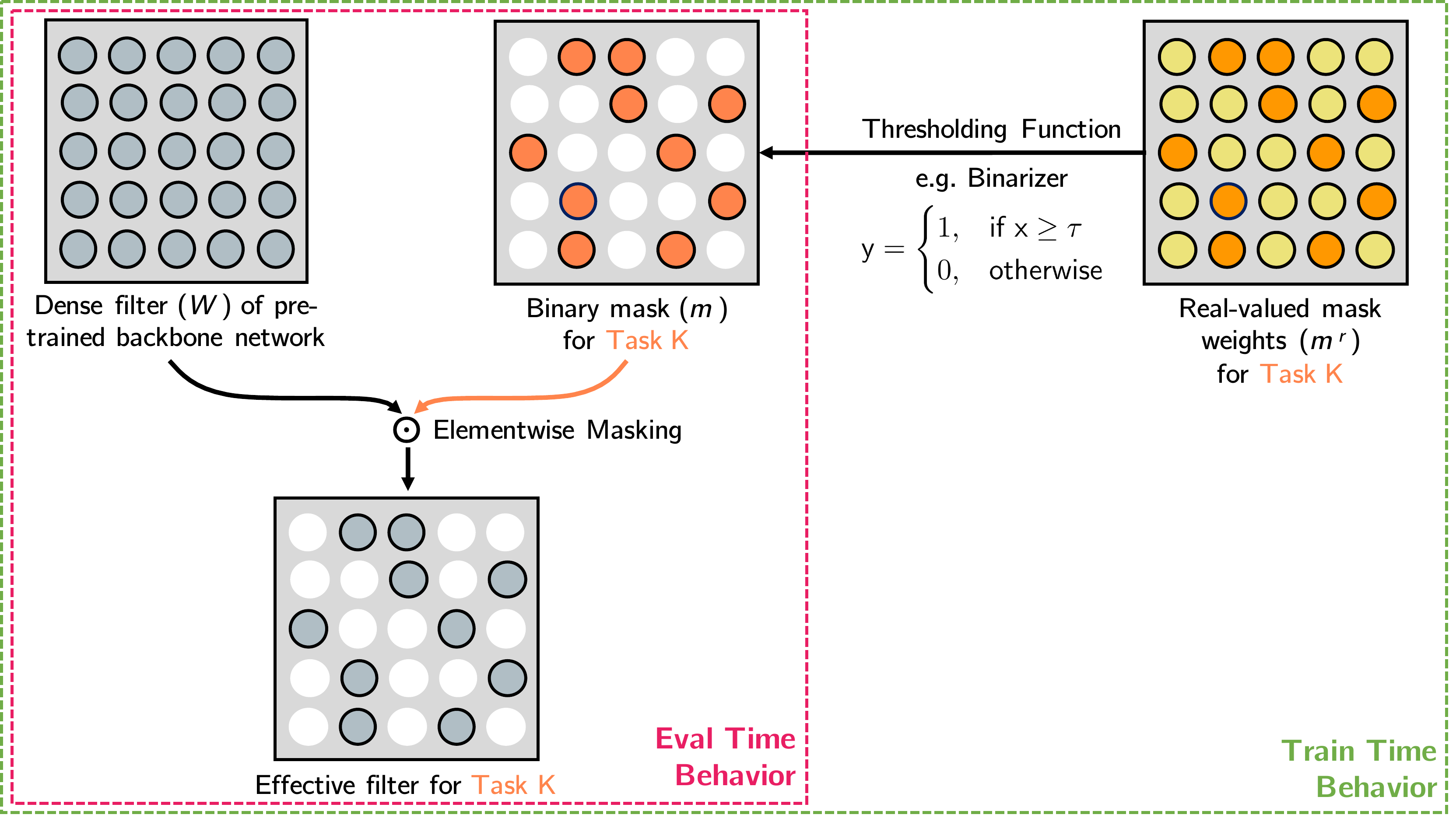}
\caption{Overview of our method for learning piggyback masks for fixed backbone networks. During training, we maintain a set of real-valued weights $m^r$ which are passed through a thresholding function to obtain binary-valued masks $m$. These masks are applied to the weights $W$ of the backbone network in an elementwise fashion, keeping individual weights active, or masked out. The gradients obtained through backpropagation of the task-specific loss are used to update the real-valued mask weights. After training, the real-valued mask weights are discarded and only the thresholded mask is retained, giving one network mask per task. 
}
\label{fig:overview}
\end{figure*}

The most popular method used in prior work for training a deep network for a new task or dataset is fine-tuning an established pre-trained model, such as the VGG-16~\cite{simonyan14VGG} trained on ImageNet classification~\cite{ILSVRC15}. A major drawback of fine-tuning is the phenomenon of \lq\lq\emph{catastrophic forgetting}\rq\rq~\cite{french1999catastrophic}, by which performance on the old task degrades significantly as the new task is learned, necessitating one to store specialized models for each task or dataset. For achieving progress towards continual learning~\cite{kirkpatrick2017overcoming,rannen2017encoder}, 
we need better methods for augmenting capabilities of an existing network 
while avoiding catastrophic forgetting and requiring as few additional parameters as possible.

Prior methods for avoiding catastrophic forgetting, such as Learning without Forgetting (LwF)~\cite{li2016learning} and Elastic Weight Consolidation (EWC)~\cite{kirkpatrick2017overcoming}, maintain performance on older tasks through proxy losses and regularization terms while modifying network weights. 
Another recent work, PackNet~\cite{mallya2017packnet}, adopts a different route of iteratively pruning unimportant weights and fine-tuning them for learning new tasks. 
As a result of pruning and weight modifications, a binary parameter usage mask is produced by PackNet.
We question whether the weights of a network have to be changed at all to learn a new task, or whether we can get by with just selectively masking, or setting certain weights to 0, while keeping the rest of the weights the same as before.
Based on this idea, we propose a novel approach in which we learn to mask weights of an existing ``\emph{backbone}'' network for obtaining good performance on a new task, as shown in Figure~\ref{fig:overview}. Binary masks that take values in $\{0,1\}$ are learned in an end-to-end differentiable fashion while optimizing for the task at hand. These masks are elementwise applied to backbone weights, allowing us to learn a large range of different filters, even with fixed weights. 
We find that a well-initialized backbone network is crucial for good performance and that the popular ImageNet pre-trained network generalizes to multiple new tasks. 
After training for a new task, we obtain a per-task binary mask that simply ``\emph{piggybacks}'' onto the backbone network.

Our experiments conducted on image classification, and presented in Section~\ref{sec:results}, show that this proposed method obtains performance similar to using a separate network per task, for a variety of datasets considered in prior work~\cite{mallya2017packnet} such as CUBS birds~\cite{WahCUB_200_2011}, Stanford cars~\cite{krause20133d}, Oxford flowers~\cite{Nilsback08}, as well 
datasets with a significant departure from the natural image domain of the ImageNet dataset such as WikiArt paintings~\cite{saleh2015large} and human sketches~\cite{eitz2012humans}. 
We demonstrate the applicability of our method to multiple network architectures including VGG-16~\cite{simonyan14VGG}, ResNets~\cite{he2016deep,zagoruyko2016wide}, and DenseNets~\cite{huang2017densely}.
Section~\ref{sec:analysis} tries to offer some insight into the workings of the proposed method, and analyzes design choices that affect performance.
As presented in Section~\ref{sec:other_results}, we also obtain performance competitive with the best methods~\cite{rosenfeld2017incremental} on the Visual Decathlon challenge~\cite{rebuffi2017learning} while using the least amount of additional parameters. Finally, we show that our method can be used to train a fully convolutional network for semantic segmentation starting from a classification backbone.


\section{Related Work}
\label{sec:related_work}


While multiple prior works~\cite{bilen2016integrated,caruana1998multitask,kokkinos2016ubernet} have explored multi-task training, wherein data of all tasks is available at the time of training, we consider the setting in which new tasks are available sequentially, a more realistic and challenging scenario.
Prior work under this setting is based on Learning without Forgetting (LwF)~\cite{rannen2017encoder,li2016learning,shmelkov2017incremental} and Elastic Weight Consolidation (EWC)~\cite{kirkpatrick2017overcoming,lee2017overcoming}. LwF uses initial network responses on new data as regularization targets during new task training, while EWC imposes a smooth penalty on changing weights deemed to be important to prior tasks. An issue with these methods is that it is not possible to determine the change in performance on prior tasks beforehand since all weights of the network are allowed to be modified to varying degrees. PackNet~\cite{mallya2017packnet} avoids this issue by identifying weights important for prior tasks through network pruning, and keeping the important weights fixed after training for a particular task. Additional information is stored per weight parameter of the network to indicate which tasks it is used by. However, for each of these methods, performance begins to drop as many tasks are added to the network. In the case of LwF, a large domain shift for a new task causes significant drop in prior task performance~\cite{li2016learning}. For PackNet, performance on a task drops as it is added later to the network due to the lack of available free parameters, and the total number of tasks that can be added is ultimately limited due to the fixed size of the network~\cite{mallya2017packnet}. 

Our proposed method does not change weights of the initial backbone network and learns a different mask per task. As a result, it is agnostic to task ordering and the addition of a task does not affect performance on any other task. Further, an unlimited number of tasks can piggyback onto a backbone network by learning a new mask. 
The parameter usage masks in PackNet were obtained as a by-product of network pruning~\cite{han2015learning}, but we learn appropriate masks based on the task at hand.
This idea of masking is related to PathNet~\cite{fernando2017pathnet}, which learns selective routing through neurons using  evolutionary strategies. We achieve similar behavior through an end-to-end differentiable method, which is less computationally demanding. The learning of separate masks per task decouples the learning of multiple tasks, freeing us from having to choose hyperparameters such as batch mixing ratios~\cite{kokkinos2016ubernet}, pruning ratios~\cite{mallya2017packnet}, and cost weighting~\cite{li2016learning}.

Similar to our proposed method, another set of methods adds new tasks by learning additional task-specific parameters. For a new task, Progressive Neural Networks~\cite{rusu2016progressive} duplicates the base architecture while adding lateral connections to layers of the existing network. The newly added parameters are optimized for the new task, while keeping old weights fixed. This method incurs a large overhead as the network is replicated for the number of tasks added. 
The method of Residual Adapters~\cite{rebuffi2017learning} develops on the observation that linearly parameterizing a convolutional filter bank of a network is the same as adding an additional per-task convolutional layer to the network. 
The most recent Deep Adaptation Networks (DAN)~\cite{rosenfeld2017incremental} allows for learning new filters that are linear combinations of existing filters. Similar to these methods, we enable the learning of new per-task filters. However, these new filters are constrained to be masked versions of existing filters.
Our learned binary masks incur an overhead of 1 bit per network parameter, smaller than all of the prior work. Further, we do not find it necessary to learn task-specific layer biases and batch normalization parameters.

Our method for training binary masks is based on the technique introduced by Courbariaux \emph{et al.}~\cite{courbariaux2015binaryconnect,hubara2016binarized} for the training of a neural network with binary-valued weights from scratch. The authors maintain a set of real-valued weights that are passed through a binarizer function during the forward pass. Gradients are computed with respect to the binarized weights during the backward pass through the application of the chain rule, and the real-valued weights are updated using the gradients computed for the binarized versions. In~\cite{courbariaux2015binaryconnect}, the authors argue that even though the gradients computed in this manner are noisy, they effectively serve as a regularizer and quantization errors cancel out over multiple iterations. Subsequent work including~\cite{li2016ternary,zhu2016trained} has extended this idea to ternary-valued weights. Unlike these works, we do not train a quantized network from scratch but instead learn quantized masks that are applied to fixed, real-valued filter weights. Work on sparsifying dense neural networks, specifically~\cite{guo2016dynamic}, has used the idea of masked weight matrices. However, only their weight matrix was trainable and their mask values were a fixed function of the magnitude of the weight matrix and not explicitly trainable. 
In contrast, we treat the weight matrix of the backbone network as a fixed constant, and our proposed approach combines the key ideas from these two areas of network binarization and masked weight matrices to learn piggyback masks for new tasks.

\section{Approach}
\label{sec:approach}

The key idea behind our method is to learn to selectively mask the fixed weights of a base network, so as to improve performance on a new task. We achieve this by maintaining a set of real-valued weights that are passed through a deterministic thresholding function to obtain binary masks, that are then applied to existing weights. By updating the real-valued weights through backpropagation, we hope to learn binary masks appropriate for the task at hand. This process is illustrated in Figure~\ref{fig:overview}.
By learning different binary-valued $\{0, 1\}$ masks per task, which are element-wise applied to network parameters, we can re-use the same underlying base network for multiple tasks, with minimal overhead.
Even though we do not modify the weights of the network, a large number of different filters can be obtained through masking. For example, a dense weight vector such as $[0.1, 0.9, -0.5, 1]$ can give rise to filters such as  $[0.1, 0, 0, 1],  [0, 0.9, -0.5, 0]$, and  $[0, 0.9, -0.5, 1]$ after binary masking. 
In practice, we begin with a network such as the VGG-16 or ResNet-50 pre-trained on the ImageNet classification task as our base network, referred to as the \emph{backbone} network, and associate a real-valued mask variable with each weight parameter of all the convolutional and fully-connected layers. By combining techniques used in network binarization~\cite{courbariaux2015binaryconnect,hubara2016binarized} and pruning~\cite{guo2016dynamic}, we train these mask variables to learn the task at hand in an end-to-end fashion, as described in detail below.
The choice of the initialization of the backbone network is crucial for obtaining good performance, and is further analyzed in Section~\ref{subsec:init_matter}.

For simplicity, we describe the mask learning procedure using the example of a fully-connected layer, but this idea can easily be extended to a convolutional layer as well.
Consider a simple fully-connected layer in a neural network. 
Let the input and output vectors be denoted by $\mathbf{x}=(x_1, x_2, \cdots, x_m)^T$ of size $m \times 1$, and $\mathbf{y}=(y_1, y_2, \cdots, y_n)^T$ of size $n \times 1$, respectively. Let the weight matrix of the layer be $\mathbf{W}=[w]_{ji}$ of size $n\times m$. The input-output relationship is then given by $\mathbf{y}=\mathbf{W}\mathbf{x}$, or $y_j=\sum_{i=1}^m w_{ji} \cdot x_i$. 
The bias term is ignored for ease of notation.
Let $\delta v$ denote the partial derivative of the error function $E$ with respect to the variable $v$.
The backpropagation equation for the weights $\mathbf{W}$ of this fully-connected layer is given by
\vspace{-4pt}
\begin{align}
\delta w_{ji} \triangleq \frac{\partial E}{\partial w_{ji}} &= \left(\frac{\partial E}{\partial y_j}\right) \cdot \left(\frac{\partial y_j}{\partial w_{ji}}\right)\\
&= \delta y_j \cdot x_i\\
\therefore \delta \mathbf{W} &\triangleq \left[ \frac{\partial E}{\partial w} \right]_{ji} = \delta\mathbf{y} \cdot \mathbf{x}^T,
\end{align}
where $\delta\mathbf{y}=(\delta y_1, \delta y_2, \cdots, \delta y_n)^T$ is of size $n \times 1$.

Our modified fully-connected layer associates a matrix of real-valued mask weights $\mathbf{m^r}=[m^r]_{ji}$ with every weight matrix $\mathbf{W}$, of the same size as $\mathbf{W}$ ($n \times m$), as indicated by the rightmost filter in Figure~\ref{fig:overview}. We obtain thresholded mask matrices $\mathbf{m}=[m]_{ji}$ by passing the real-valued mask weight matrices $\mathbf{m^r}$ through a hard binary thresholding function given by
\vspace{-4pt}
\begin{equation}
  {m}_{ji} = 
    \begin{cases}
      1, & \text{if } {m^r}_{ji} \geq \tau\\
      0, & \text{otherwise}
    \end{cases},
    \label{eq:tau}
\end{equation}
where $\tau$ is a selected threshold.
The binary-valued matrix $\mathbf{m}$ activates or switches off contents of $\mathbf{W}$ depending on whether a particular value $m_{ji}$ is 0 or 1. The layer's input-output relationship is given by the equation $\mathbf{y} = (\mathbf{W} \odot \mathbf{m})\,\mathbf{x}$, or $y_j=\sum_{i=1}^m w_{ji} \cdot m_{ji} \cdot x_i$, where $\odot$ indicates elementwise multiplication or masking. As mentioned previously, we set the weights $\mathbf{W}$ of our modified layer to those from the same architecture pre-trained on a task such as ImageNet classification.
We treat the weights $\mathbf{W}$ as fixed constants throughout, while only training the real-valued mask weights $\mathbf{m^r}$.
The backpropagation equation for the thresholded mask weights $\mathbf{m}$ of this fully-connected layer is given by
\begin{align}
\delta m_{ji} \triangleq \frac{\partial E}{\partial m_{ji}} &= \left(\frac{\partial E}{\partial y_j}\right) \cdot \left(\frac{\partial y_j}{\partial m_{ji}}\right)\\
&= \delta y_j \cdot w_{ji} \cdot x_i\\
\therefore \delta \mathbf{m} &\triangleq \left[ \frac{\partial E}{\partial m} \right]_{ji} = (\delta\mathbf{y} \cdot \mathbf{x}^T) \odot \mathbf{W}.
\label{eq:dmask}
\end{align}

Even though the hard thresholding function is non-differentiable, the gradients of the thresholded mask values $\mathbf{m}$ serve as a noisy estimator of the gradients of the real-valued mask weights $\mathbf{m^r}$, and can even serve as a regularizer, as shown in prior work~\cite{courbariaux2015binaryconnect,hubara2016binarized}. We thus update the real-valued mask weights $\mathbf{m^r}$ using gradients computed for $\mathbf{m}$, the thresholded mask values. After adding a new final classification layer for the new task, the entire system can be trained in an end-to-end differentiable manner. In our experiments, we did not train per-task biases as prior work~\cite{mallya2017packnet} showed that this does not have any significant impact on performance. We also did not train per-task batch-normalization parameters for simplicity. Section~\ref{subsec:training_bn} analyzes the benefit of training per-task batchnorm parameters, especially for tasks with large domain shifts.

After training a mask for a given task, we no longer require the real-valued mask weights. They are discarded, and only the thresholded masks associated with the backbone network layers are stored. A typical neural network parameter is represented using a 32-bit float value (including in our PyTorch implementation). A binary mask only requires 1 extra bit per parameter, leading to an approximate per-task overhead of 1/32 or 3.12\% of the backbone network size.  

\medskip
\noindent{\bf Practical optimization details.} From Eq.~\ref{eq:dmask}, we observe that $|\delta\mathbf{m}|, |\delta\mathbf{m^r}|\propto|\mathbf{W}|$. The magnitude of pre-trained weights varies across layers of a network, and as a result, the mask gradients would also have different magnitudes at different layers.
This relationship requires us to be careful about the manner in which we initialize and train mask weights $\mathbf{m^r}$. There are two possible approaches:

\noindent 1) Initialize $\mathbf{m^r}$ with values proportional to the weight matrix $\mathbf{W}$ of the corresponding layer. In this case, the ratio 
$|\delta\mathbf{m^r}|/|\mathbf{m^r}|$
will be similar across layers, and a constant learning rate can be used for all layers.

\noindent 2) Initialize $\mathbf{m^r}$ with a constant value, such as $0.01$, for all layers. This would require a separate learning rate per layer, due to the scaling of the mask gradient by the layer weight magnitude.
While using SGD, scaling gradients obtained at each layer by a factor of $1/\text{avg}(|\mathbf{W}|)$, while using a constant learning rate, has the same effect as layer-dependent learning rates. Alternatively, one could use adaptive optimizers such as Adam, which would learn appropriate scaling factors.

The second initialization approach combined with the Adam optimizer produced the best results, with a consistent gain in accuracy by $\sim$ 2\% compared to the alternatives.
We initialized the real-valued weights with a value of 1e-2 with a binarizer threshold ($\tau$, in Equation~\ref{eq:tau}) of 5e-3 in all our experiments. Randomly initializing the real-valued mask weights such that the thresholded binary masks had an equal number of 0s and 1s did not give very good performance. Ensuring that all thresholded mask values were 1 provides the same network initialization as that of the baseline methods.

We also tried learning ternary masks $\{-1, 0, 1\}$ by using a modified version of Equation~\ref{eq:tau} with two cut-off thresholds, but did not achieve results that were significantly different from those obtained with binary masks. As a result, we only focus on results obtained with binary masks in the rest of this work.



\section{Experiments and Results}
\label{sec:results}
\label{subsec:main_results}

\begin{figure*}
  \begin{minipage}{0.55\textwidth}
    \centering
    \vspace{9pt}
    \resizebox{\columnwidth}{!}{%
      \begin{tabular}{l|c|c|c}
        \hline
        {\bf Dataset} & {\bf \#Train} & {\bf \#Eval} & {\bf \#Classes} \\\hline\hline
        ImageNet~\cite{ILSVRC15} & 1,281,144 & 50,000 & 1,000 \\\hline
        Places365~\cite{zhou2017places} & 1,803,460 & 36,500 & 365 \\\hline
        CUBS~\cite{WahCUB_200_2011} & 5,994 & 5,794 & 200 \\\hline
        Stanford Cars~\cite{krause20133d} & 8,144 & 8,041 & 196 \\\hline
        Flowers~\cite{Nilsback08} & 2,040 & 6,149 & 102 \\\hline
        WikiArt~\cite{saleh2015large} & 42,129 & 10,628 & 195 \\\hline
        Sketch~\cite{eitz2012humans} & 16,000 & 4,000 & 250 \\\hline
      \end{tabular}
      } 
    \captionof{table}{Summary of datasets used.}
    \label{table:dataset_stats}
  \end{minipage}
  \hfill
  \begin{minipage}{.42\textwidth}
    \centering
    \subfloat[WikiArt]{\includegraphics[trim={1.2cm 3cm 0.4cm 0.4cm}, clip, width=0.47\linewidth]{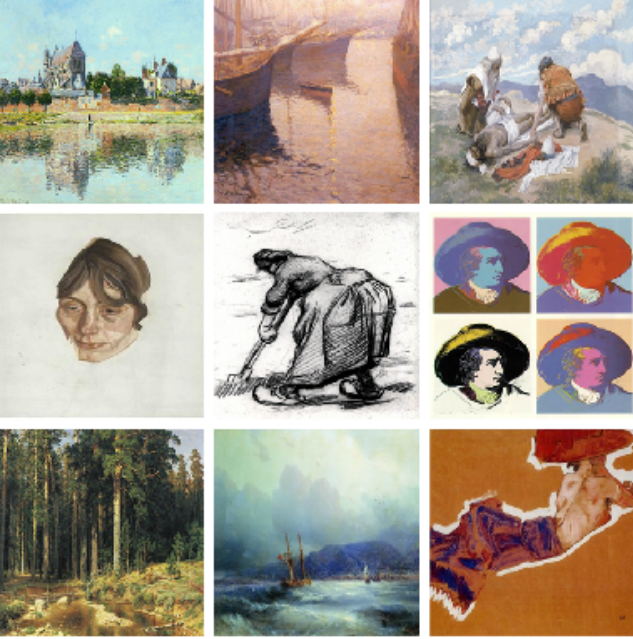}\label{fig:wikiart}}
    \hfill
    \subfloat[Sketch]{\includegraphics[trim={1.2cm 3cm 0.4cm 0.4cm}, clip, width=0.47\linewidth]{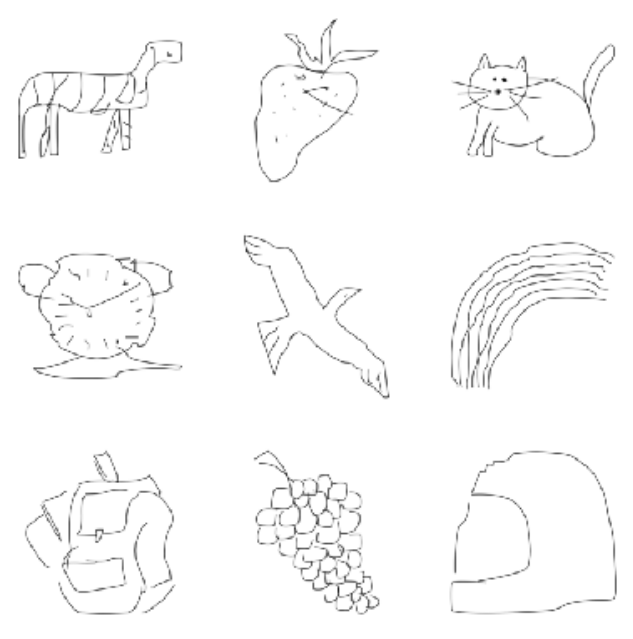}\label{fig:sketch}}
    \caption{Datasets unlike ImageNet.}
    \label{fig:dataset}
  \end{minipage}
\end{figure*}

We consider a wide variety of datasets, statistics of which are summarized in Table~\ref{table:dataset_stats}, to evaluate our proposed method.
Similar to PackNet~\cite{mallya2017packnet}, we evaluate our method on two large-scale datasets, the ImageNet object classification dataset~\cite{ILSVRC15} and the Places365 scene classification dataset~\cite{zhou2017places}, each of which has over a million images, as well as the CUBS~\cite{WahCUB_200_2011}, Stanford Cars~\cite{krause20133d}, and Flowers~\cite{Nilsback08} fine-grained classification datasets. Further, we include two more datasets with significant domain shifts from the natural images of ImageNet, the WikiArt Artists classification dataset, created from the WikiArt dataset~\cite{saleh2015large}, and the Sketch classifcation dataset~\cite{eitz2012humans}. The former includes a wide genre of painting styles, as shown in Figure~\ref{fig:wikiart}, while the latter includes black-and-white sketches drawn by humans, as shown in Figure~\ref{fig:sketch}. For all these datasets, we use networks with an input image size of $224\times224$ px.

Table~\ref{table:results_piggyback} reports the errors obtained on fine-grained classification tasks by learning binary-valued piggyback masks for a VGG-16 network pre-trained on ImageNet classification. 
The first baseline considered is {\bf Classifier Only}, which only trains a linear classifier using \texttt{fc7} features extracted from the pre-trained VGG-16 network. This is a commonly used method that has low overhead as all layers except for the last classification layer are re-used amongst tasks.
The second and more powerful baseline is {\bf Individual Networks}, which finetunes a separate network per task. We also compare our method to the recently introduced {\bf PackNet}~\cite{mallya2017packnet} method, which adds multiple tasks to a network through iterative pruning and re-training. 
We train all methods for 30 epochs. We train the piggyback and classifier only, using the Adam optimizer with an initial learning rate of 1e-4, which is decayed by a factor of 10 after 15 epochs. We found SGDm with an initial learning rate of 1e-3 to work better for the individual VGG network baseline.
For PackNet, we used a 50\% pruned initial network trained with SGDm with an initial learning rate of 1e-3 using the same decay scheme as before.
We prune the network by 75\% and re-train for 15 epochs with a learning rate of 1e-4 after each new task is added. All errors are averaged over 3 independent runs.

\begin{table*}[th!]
  \centering
  \begin{tabular}{l||c|c|c|c|c}
    \hline
    \multirow{2}{*}{\bf Dataset} & {\bf Classifier} & \multicolumn{2}{c|}{\bf PackNet~\cite{mallya2017packnet}} & {\bf Piggyback} & {\bf Individual} \\
    & {\bf Only} & $\downarrow$ & $\uparrow$ & {\bf (ours)}& {\bf Networks} \\\hline\hline
    \multirow{2}{*}{ImageNet} & 28.42 & \multicolumn{2}{c|}{29.33} & 28.42 & 28.42 \\
    & (9.61) & \multicolumn{2}{c|}{(9.99)} & (9.61) & (9.61) \\\hline
    CUBS & 36.49 & 22.30 & 29.69 & 20.99 & 21.30 \\\hline
    Stanford Cars & 54.66 & 15.81 & 21.66 & 11.87 & 12.49 \\\hline
    Flowers & 20.01 & 10.33 & 10.25 & 7.19 & 7.35 \\\hline
    WikiArt & 49.53 & 32.80 & 31.48 & 29.91 & 29.84 \\\hline
    Sketch & 58.53 & 28.62 & 24.88 & 22.70 & 23.54 \\\hline\hline
    \# Models (Size) & 1 (537 MB) & \multicolumn{2}{c|}{1 (587 MB)} & 1 (621 MB) & 6 (3,222 MB) \\\hline
  \end{tabular}
  \caption{Errors obtained by starting from an ImageNet-trained VGG-16 network and then using various methods to learn new fine-grained classification tasks. PackNet performance is sensitive to order of task addition, while the rest, including our proposed method, are agnostic. $\downarrow$ and $\uparrow$ indicate that tasks were added in the CUBS $\rightarrow$ Sketch, and Sketch $\rightarrow$ CUBS order, resp. 
  Values in parentheses are top-5 errors, rest are top-1 errors.}
  \label{table:results_piggyback}
\end{table*}

As seen in Table~\ref{table:results_piggyback}, training individual networks per task clearly provides a huge benefit over the classifier only baseline for all tasks. 
PackNet significantly improves over the classifier only baseline, but begins to suffer when more than 3 tasks are added to a single network. As PackNet is sensitive to the ordering of tasks, we try two settings - adding tasks in order from CUBS to Sketch (top to bottom in Table~\ref{table:results_piggyback}), and the reverse.
The order of new task addition has a large impact on the performance of PackNet, with errors increasing by 4-7\% as the addition of a task is delayed from first to last (fifth). The error on ImageNet is also higher in the case of PackNet, due to initial network pruning. 
By training binary piggyback masks, we are able to obtain errors slightly lower than the individual network case. We believe that this is due to the regularization effect caused by the constrained filter modification allowed by our method. Due to the learning of independent masks per task, the obtained performance is agnostic to the ordering of new tasks, albeit at a slightly higher storage overhead as compared to PackNet. The number of weights switched off varies per layer and by dataset depending on its similarity to the ImageNet dataset. This effect is further examined in Section~\ref{subsec:sparsity_distribution}.

\begin{table*}[t!]
  \centering
  \begin{tabular}{l||c|c|c|c}
    \hline
    \multirow{2}{*}{\bf Dataset} & {\bf Jointly Trained} & {\bf PackNet} & {\bf Piggyback} & {\bf Individual} \\ 
    & {\bf Network$^\ast$} & {\bf \cite{mallya2017packnet}} & {\bf (ours)} & {\bf Networks} \\\hline\hline
    \multirow{2}{*}{ImageNet} & 33.49 & 29.33 & 28.42 & 28.42 \\
    & (12.25) & (9.99) & (9.61) & (9.61) \\\hline
    \multirow{2}{*}{Places365} & 45.98 & 46.64 & 46.71 & 46.35 \\
    & (15.59) & (15.92) & (16.18) & (16.14)$^\ast$ \\\hline\hline
    \# Models (Size) & 1 (537 MB) & {1 (554 MB)} & 1 (554 MB) & {2 (1,074 MB)} \\\hline
  \end{tabular}
  \caption{Adding a large-scale dataset to an ImageNet-trained VGG-16 network. Values in parentheses are top-5 errors, rest are top-1 errors. 
  $^\ast$ indicates models downloaded from {\small \url{https://github.com/CSAILVision/places365}}, trained by~\cite{zhou2017places}.}
  \label{table:results_places}
\end{table*}

While the results above were obtained by adding multiple smaller fine-grained classification tasks to a network, the next set of results in Table~\ref{table:results_places} examines the effect of adding a large-scale dataset, the Places365 scene classification task with 1.8M images, to a network.
Here, instead of the Classifier Only baseline, we compare against the {\bf Jointly Trained Network} of~\cite{zhou2017places}, in which a single network is simultaneously trained for both tasks. 
Both PackNet and Piggyback were trained for 20 epochs on Places365.
Once again, we are able to achieve close to best-case performance on the Places365 task, obtaining top-1 errors within 0.36\% of the individual network, even though the baselines were trained for 60-90 epochs~\cite{zhou2017places}. The performance is comparable to PackNet, and for the case of adding just one task, both incur a similar overhead.

\begin{table*}[th!]
  \centering
  \begin{tabular}{l||c|c|c|c|c}
    \hline
    \multirow{2}{*}{\bf Dataset} & {\bf Classifier} & \multicolumn{2}{c|}{\bf PackNet~\cite{mallya2017packnet}} & {\bf Piggyback} & {\bf Individual} \\
    & {\bf Only} & $\downarrow$ & $\uparrow$ & {\bf (ours)}& {\bf Networks} \\\hline\hline
    \multicolumn{6}{c}{\bf VGG-16 BN} \\\hline
    \multirow{2}{*}{ImageNet} & 26.63 & \multicolumn{2}{c|}{27.18} & 26.63& 26.63\\
    & (8.49) & \multicolumn{2}{c|}{(8.69)} & (8.49) & (8.49)\\\hline
    CUBS & 33.88 & 20.21 & 23.82 & 18.37 & 19.57 \\\hline
    Stanford Cars & 51.62 & 14.05 & 17.60 & 9.87 & 9.41 \\\hline
    Flowers & 19.38 & 7.82 & 7.85 & 4.84 & 4.55 \\\hline
    WikiArt & 48.05 & 30.21 & 29.59 & 27.50 & 26.68 \\\hline
    Sketch & 59.96 & 25.47 & 23.53 & 21.41 & 21.92 \\\hline\hline
    \# Models (Size) & 1 (537 MB) & \multicolumn{2}{c|}{1 (587 MB)} & 1 (621 MB) & 6 (3,222 MB)\\\hline
    \multicolumn{6}{c}{\bf ResNet-50} \\\hline
    \multirow{2}{*}{ImageNet} & 23.84 & \multicolumn{2}{c|}{24.29} & 23.84 & 23.84\\
    & (7.13) & \multicolumn{2}{c|}{(7.18)} & (7.13) & (7.13)\\\hline
    CUBS & 29.97 & 19.59 & 28.62 & 18.41 & 17.17 \\\hline
    Stanford Cars & 47.20 & 13.89 & 19.99 & 10.38 & 8.17 \\\hline
    Flowers & 14.01 & 6.96 & 9.45 & 5.23 & 3.44 \\\hline
    WikiArt & 44.40 & 30.60 & 29.69 & 28.67 & 24.40 \\\hline
    Sketch & 49.14 & 23.83 & 21.30 & 20.09 & 19.22 \\\hline\hline
    \# Models (Size) & 1 (94 MB) & \multicolumn{2}{c|}{1 (103 MB)} & 1 (109 MB) & 6 (564 MB)\\\hline
    \multicolumn{6}{c}{\bf DenseNet-121} \\\hline
    \multirow{2}{*}{ImageNet} & 25.56 & \multicolumn{2}{c|}{25.60} & 25.56 & 25.56\\
    & (8.02) & \multicolumn{2}{c|}{(7.89)} & (8.02) & (8.02) \\\hline
    CUBS & 26.55 & 19.26 & 30.36 & 19.50 & 18.08 \\\hline
    Stanford Cars & 43.19 & 15.35 & 22.09 & 10.87 & 8.64 \\\hline
    Flowers & 16.56 & 8.94 & 8.46 & 5.31 & 3.49 \\\hline
    WikiArt & 45.08 & 33.66 & 30.81 & 29.56 & 23.59 \\\hline
    Sketch & 46.88 & 25.35 & 21.08 & 20.30 & 19.48 \\\hline\hline
    \# Models (Size) & 1 (28 MB) & \multicolumn{2}{c|}{1 (31 MB)} & 1 (33 MB) & 6 (168 MB)\\\hline
  \end{tabular}
  \caption{Results on other network architectures. Values in parentheses are top-5 errors, rest are top-1 errors. $\uparrow$ and $\downarrow$ indicate order of task addition for PackNet.}
  \label{table:results_piggyback_other_networks}
\end{table*}

The previous results were obtained using the large VGG-16 network, and it is not immediately obvious whether the piggyback method would work for much deeper networks that have batch normalization layers. Masking out filter weights can change the average magnitude of activations, requiring changes to batchnorm parameters. 
We present results obtained with a VGG-16 network with batch normalization layers, the ResNet-50, and DenseNet-121 networks in Table~\ref{table:results_piggyback_other_networks}.
We observe that the method can be applied without any changes to these network architectures with batchnorm, residual, and skip connections. In the presented results, we do not learn task-specific batchnorm parameters.
We however notice that the deeper a network gets, the larger the gap between the performance of piggyback and individual networks. For the VGG-16 architecture, piggyback can often do as well as or better than individual models, but for the ResNet and DenseNet architectures, the gap is $\sim$2\%.
In Section~\ref{subsec:training_bn} we show that learning task-specific batchnorm parameters in the case of datasets that exhibit a large domain shift, such as WikiArt, for which the performance gap is $4$-$5\%$ (as seen in Table~\ref{table:results_piggyback_other_networks}), helps further close the gap. 


\section{Analysis}
\label{sec:analysis}

\subsection{Does Initialization Matter?}
\label{subsec:init_matter}
Here, we analyze the importance of the initialization of the backbone network. It is well known that training a large network such as the VGG-16 from scratch on a small dataset such as CUBS, or Flowers leads to poor performance, and the most popular approach is to fine-tune a network pre-trained on the ImageNet classification task. It is not obvious whether initialization is just as important for the piggyback method. Table~\ref{table:diff_inits} presents the errors obtained by training piggyback masks for tasks using the ResNet-50 as the backbone network, but with different initializations. We consider 3 different initializations: 1) a network trained on the ImageNet classification task, the popular initialization for fine-tuning, 2) a network trained from scratch on the Places365 scene classification task, a dataset larger than ImageNet (1.8 M v/s 1.3 M images), but with fewer classes (365 v/s 1000), and lastly 3) a randomly initialized network.

We observe in Table~\ref{table:diff_inits} that initialization does indeed matter, with the ImageNet-initialized network outperforming both the Places365 and randomly initialized network on all tasks. In fact, by training a piggyback mask for the Places365 dataset on an ImageNet-initialized backbone network, we obtain an accuracy very similar to a network trained from scratch on the Places365 dataset.
The ImageNet dataset is very diverse, with classes ranging from animals, to plants, cars and other inanimate objects, whereas the Places365 dataset is solely devoted to the classification of scenes such as beaches, bedrooms, restaurants, \emph{etc}. As a result, the features of the ImageNet-trained network serve as a very general and flexible initialization
A very interesting observation is that even a randomly initialized network obtains non-trivial accuracies on all datasets. This indicates the learning a mask is indeed a powerful technique of utilizing fixed filters and weights for adapting a network to a new task.

\begin{table*}[h!]
  \centering
  \begin{tabular}{l||c|c|c}
    \hline
    \multirow{2}{*}{\bf Dataset} & \multicolumn{3}{c}{\bf Pre-training/Initialization}\\
     & {\bf ImageNet} & {\bf Places365} & {\bf Random}\\\hline\hline
    CUBS & 18.41 & 28.50 & 66.24 \\\hline
    Stanford Cars & 10.38 & 13.70 & 77.79 \\\hline
    Flowers & 5.23 & 10.92 & 71.17 \\\hline
    WikiArt & 28.67 & 31.24 & 64.74 \\\hline
    Sketch & 20.09 & 23.17 & 43.75 \\\hline
    \multirow{2}{*}{ImageNet} & {23.84} & 32.56 & 71.48 \\
    & ({7.13}) & (11.92) & (46.73) \\\hline
    \multirow{2}{*}{Places365} & 45.17 & {45.39} & 60.41 \\
    & (15.12) & ({15.05}) & (28.94) \\\hline
  \end{tabular}
  \caption{Errors obtained by piggyback masks for the ResNet-50 backbone network with different initializations. Errors in parentheses are top-5 errors,  the rest are top-1 errors.}
  \label{table:diff_inits}
\end{table*}

\subsection{Learned sparsity and its distribution across network layers}
\label{subsec:sparsity_distribution}
Table~\ref{table:perc_zero} reports the total sparsity, or the number of mask values set to 0 in a binary piggyback mask learned for the corresponding choice of dataset and network architecture. This measures the amount of change that is required to be made to the backbone network, or the deviation from the ImageNet pre-trained initialization, in order to obtain good performance on a given dataset.
We note that the amount of sparsity obtained on fine-grained datasets seems to be proportional to the errors obtained by the Classifier Only method on the respective datasets. The easiest Flowers dataset requires the least number of changes, or a sparsity of $4.51\%$, while the harder WikiArt dataset leads to a $34.14\%$ sparsity for a VGG-16 network mask. Across network architectures, we observe a similar pattern of sparsity based on the difficulty of the tasks. The sparsity obtained is also a function of the magnitude of the real-valued mask initialization and threshold used for the binarization (See Equation~\ref{eq:tau}), with a higher threshold leading to higher sparsity. The numbers in Table~\ref{table:perc_zero}
were obtained using our default settings of a binarizer threshold of 5e-3 and a uniform real-valued mask initialization of 1e-2.

\begin{table*}[b!]
  \centering
  \begin{tabular}{l||c|c|c|c|c}
    \hline
    \multirow{2}{*}{\bf Dataset} & \multirow{2}{*}{\bf VGG-16} & {\bf VGG-16} & \multicolumn{2}{c|}{\bf ResNet-50} & {\bf Dense-} \\
    \cline{4-5}
    & & {\bf BN} & {\bf ImNet-init.} & {\bf Places-init.} & {\bf Net-121}\\\hline\hline
    CUBS & 14.09\% & 13.24\% & 12.21\% & 15.22\% & 12.01\% \\\hline
    Stanford Cars & 17.03\% & 16.70\% & 15.65\% & 17.72\% & 15.80\% \\\hline
    Flowers & 4.51\% &4.52\% & 4.48\% & 6.45\% & 5.28\% \\\hline
    WikiArt & 34.14\% & 33.01\% & 30.47\% & 30.04\% & 29.11\% \\\hline
    Sketch & 27.23\% & 26.05\% & 23.04\% & 24.23\% & 22.24\% \\\hline
    ImageNet & -- & -- & -- & 37.59\% & -- \\\hline
    Places365 & 43.47\% & -- & 37.99\%  & -- & -- \\\hline
  \end{tabular}
  \caption{Percentage of zeroed out weights after training a binary mask for the respective network architectures and datasets.
  }
  \label{table:perc_zero}
\end{table*}

\begin{figure*}[h!]
  \centering
  \includegraphics[width=\textwidth, clip, trim={0cm 9cm 0cm 0cm}]{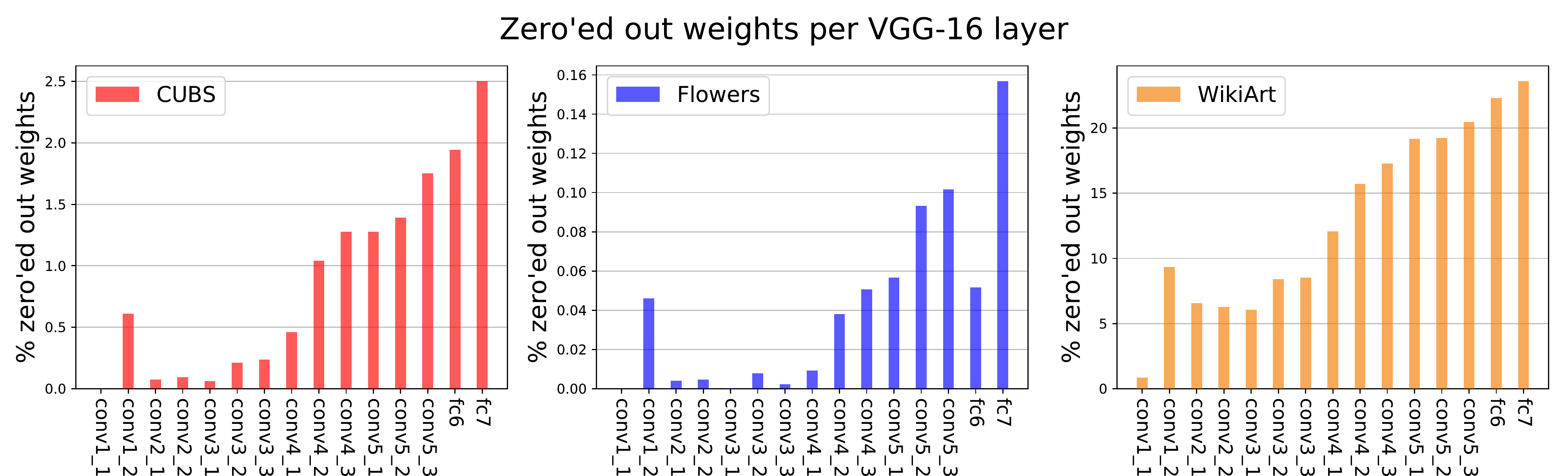}
  \caption{Percentage of weights masked out per ImageNet pre-trained VGG-16 layer. Datasets similar to ImageNet share a lot of the lower layers, and require fewer changes. The number of masked out weights increases with depth of layer.}
  \label{fig:zeros_by_layer}
\end{figure*}

We observe that a Places365-initialized network requires more changes as compared to an ImageNet-initialized network (refer to the ResNet-50 column of Table~\ref{table:perc_zero}). This once again indicates that features learned on ImageNet are more diverse and serve as better initialization than those learned on Places365.

Figure~\ref{fig:zeros_by_layer} shows the sparsity obtained per layer of the ImageNet pre-trained VGG-16 network, for three datasets considered. While the total amount of sparsity obtained per dataset is different, we observe a consistent pattern of sparsity across the layers. In general, the number of changes increases with depth of the network layer. For datasets similar to ImageNet, such as CUBS, and Flowers, we observe that the low-level features (\texttt{conv1}-\texttt{conv3}) are mostly re-used without any major changes. WikiArt, which has a significant domain shift from ImageNet, requires some changes in the low-level features. All tasks seem to require changes to the mid-level (\texttt{conv4}-\texttt{conv5}) and high-level features (\texttt{fc6}-\texttt{fc7}) in order to learn new task-specific features. Similar behavior was also observed for the deeper ResNet and DenseNet networks.


\subsection{Handling large input domain shifts}
\label{subsec:training_bn}

\begin{table}[b!]
  \centering
  \begin{tabular}{l||c|c|c}
    \hline
    \multirow{2}{*}{\bf Dataset} & \multicolumn{2}{c|}{\bf Piggyback (ours)} & {\bf Individual} \\
    \cline{2-3}
    & {\bf Fixed BN} & {\bf Trained BN} & {\bf Network} \\\hline\hline
    \multicolumn{4}{c}{\bf ResNet-50} \\\hline
    WikiArt & 28.67 & 25.92 & 24.40 \\\hline
    Sketch & 20.09 & 19.82 & 19.22 \\\hline\hline
    \multicolumn{4}{c}{\bf DenseNet-121} \\\hline
    WikiArt & 29.56 & 25.90 & 23.59 \\\hline
    Sketch & 20.30 & 20.12 & 19.48 \\\hline
  \end{tabular}
  \caption{Effect of task-specific batch normalization layers on the top-1 error.}
  \label{table:batchnorm}
\end{table}

In Table~\ref{table:results_piggyback_other_networks}, we observe that WikiArt, which has a large domain shift from the ImageNet dataset on which the backbone network was trained on, has a larger gap in performance ($4$--$5\%$) between the piggyback and individual network methods, especially for the deeper ResNet and DenseNet networks. 
Those numbers are duplicated in the Piggyback - Fixed BN and Individual Network columns of Table~\ref{table:batchnorm}. We suspect that keeping batchnorm parameters fixed while training the piggyback masks might be a reason for the gap in performance, as the domain shift is likely to cause a larger discrepancy between the ideal batchnorm parameter values and those inherited from ImageNet, the effect of which is cascaded through the large number of layers. We performed these experiments again, but while updating batchnorm parameters, and report the results in the Piggyback - Trained BN column of Table~\ref{table:batchnorm}.
The top-1 error on WikiArt reduces from 28.67\% to 25.92\% for the ResNet-50 network, and from 29.56\% to 25.90\% for the DenseNet-121 network if the batchnorm parameters are allowed to update. For the Sketch dataset, training separate batchnorm parameters leads to a small decrease in error. Task-specific batchnorm parameters thus help improve performance, while causing a small increase of $\sim$1 MB in the storage overhead for both networks considered.

\section{Results on Visual Decathlon \& Semantic Segmentation}
\label{sec:other_results}
\label{subsec:visual_decathlon}

We also evaluate our proposed method on the newly introduced Visual Decathlon challenge~\cite{rebuffi2017learning} consisting of 10 classification tasks. 
While the images of this task are of a lower resolution (72 $\times$ 72 px), they contain a wide variety of tasks such as pedestrian, digit, aircraft, and action classification, making it perfect for testing the generalization abilities of our method.
Evaluation on this challenge reports per-task accuracies, and assigns a cumulative score with a maximum value of 10,000 (1,000 per task) based on the per-task accuracies. The goal is to learn models for maximizing the total score over the 10 tasks while using the least number of parameters. 
Complete details about the challenge settings, evaluation, and datasets used can be found at {\small \url{http://www.robots.ox.ac.uk/~vgg/decathlon/}}.

\begin{table*}[h!]
  \centering
  \resizebox{\textwidth}{!}{%
  \begin{tabular}{l|c||c|c|c|c|c|c|c|c|c|c||c|c}
  \hline
    {\bf Method} & {\bf \#par} & {\bf ImNet.} & {\bf Airc.} & {\bf C100} & {\bf DPed} & {\bf DTD} & {\bf GTSR} & {\bf Flwr} & {\bf Oglt} & {\bf SVHN} & {\bf UCF} & {\bf Mean} & {\bf Score} \\\hline\hline
    Scratch~\cite{rebuffi2017learning} & 10 & 59.87 & 57.1 & 75.73 & 91.2 & 37.77 & 96.55 & 56.3 & 88.74 & 96.63 & 43.27 & 70.32 & 1625 \\\hline
    Feature~\cite{rebuffi2017learning} & 1 & 59.67 & 23.31 & 63.11 & 80.33 & 45.37 & 68.16 & 73.69 & 58.79 & 43.54 & 26.8 & 54.28 & 544 \\\hline
    Finetune~\cite{rebuffi2017learning} & 10 & 59.87 & 60.34 & 82.12 & 92.82 & 55.53 & 97.53 & 81.41 & 87.69 & 96.55 & 51.2 & 76.51 & 2500 \\\hline
    Res.\,Adapt.~\cite{rebuffi2017learning} & 2 & 59.67 & 56.68 & 81.2 & 93.88 & 50.85 & 97.05 & 66.24 & 89.62 & 96.13 & 47.45 & 73.88 & 2118 \\\hline
    Res.\,Adapt.\,(J)~\cite{rebuffi2017learning} & 2 & 59.23 & 63.73 & 81.31 & 93.3 & 57.02 & 97.47 & 83.43 & 89.82 & 96.17 & 50.28 & 77.17 & 2643 \\\hline
    DAN~\cite{rosenfeld2017incremental} & 2.17 & 57.74 & 64.12 & 80.07 & 91.3 & 56.54 & 98.46 & 86.05 & 89.67 & 96.77 & 49.38 & 77.01 & 2851 \\\hline\hline
    Piggyback (Ours) & 1.28 & 57.69 & 65.29 & 79.87 & 96.99 & 57.45 & 97.27 & 79.09 & 87.63 & 97.24 & 47.48 & 76.60 & 2838 \\\hline
  \end{tabular}
  }
  \caption{Top-1 accuracies obtained on the Visual Decathlon online test set.}
  \label{table:visual_decathlon}
\end{table*}
Table~\ref{table:visual_decathlon} reports the results obtained on the online test set of the challenge. Consistent with prior work~\cite{rebuffi2017learning,rosenfeld2017incremental}, we use a Wide Residual Network~\cite{zagoruyko2016wide} with a depth of 28, widening factor of 4, and a stride of 2 in the first convolutional layer of each block. We use the $64\times 64$ px ImageNet-trained network of~\cite{rosenfeld2017incremental} as our backbone network, and train piggyback masks for the remaining 9 datasets. We train for a total of 60 epochs per dataset, with learning rate decay by a factor of 10 after 45 epochs. The base learning rate for final classifier layer which uses SGDm was chosen from \{1e-2,  1e-3\} using cross-validation over the validation set. Adam with a base learning rate of 1e-4 was used for updating the real-valued piggyback masks. Data augmentation by random cropping, horizontal flipping, and resizing the entire image was chosen based on cross-validation. 

As observed in Table~\ref{table:visual_decathlon}, our method obtains performance competitive with the state-of-the-art, while using the least amount of additional parameters over a single network. Assuming that the base network uses 32-bit parameters, it accounts for a parameter cost of $32n$ bits, where $n$ is the number of parameters. A binary mask per dataset requires $n$ bits, leading to a total cost of approximately $(32n + 9n)=41n$ bits, or a parameter ratio of $(41/32)=1.28$, as reported.

\begin{wrapfigure}{r}{0.45\textwidth}
  \begin{center}
    \includegraphics[trim={0.2cm, 0, 0.5cm, 1cm},width=0.42\textwidth]{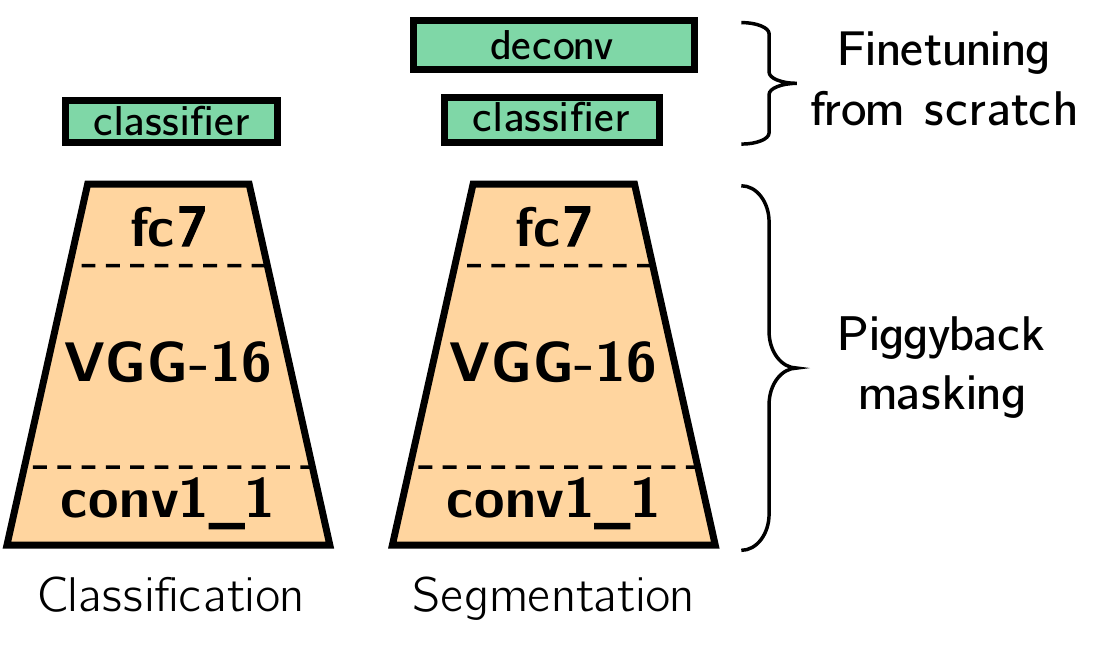}
  \end{center}
  \caption{Mixed training of layers using finetuning from scratch and piggyback masking.}
  \label{fig:mixed_training}
\end{wrapfigure}
The results presented in Section~\ref{subsec:main_results} only required a single fully connected layer to be added on top of the backbone network. Our method can also be extended to cases where more than one layers are added and trained from scratch on top of a backbone network, as shown in Figure~\ref{fig:mixed_training}.
We tested our method on the task of pixelwise segmentation using the basic Fully Convolutional Network architecture~\cite{long2015fully} which has fully connected layer followed by a deconvolutional layer of stride 32. 
We trained our networks on the 21-class PASCAL 2011 + SBD dataset, using the official splits provided by~\cite{SegmentationSplits} for 15 epochs.
Using the VGG-16 finetuned network, we obtain a mean IOU of 61.08\footnote{This is  lower than the 63.6 mIOU obtained by~\cite{long2015fully} owing to differences in the Caffe and PyTorch VGG-16 initializations, as documented at \url{https://goo.gl/quvmm2}.}.
Using the piggyback method, we obtain a competitive mean IOU of 61.41. 
Instead of replicating the whole VGG-16 network of $\sim$500 MB, we only need an overhead of 17 MB for masking the backbone network and 7.5 MB for the newly added layers.
These results show that the proposed method does not face any issues due to mixed training schemes and that piggyback masking is a competitive alternative to full-network finetuning.

\section{Conclusions}
\label{sec:conclusions}
We have presented a novel method for utilizing the fixed weights of a network for obtaining good performance on a new task, empirically showing that the proposed method works for multiple datasets and network architectures.
We hope that the piggyback method will be useful in  practical scenarios where new skills need to be learned on a deployed device without having to modify existing weights or download a new large network. The re-usability of the backbone network and learned masks should help simplify and scale the learning of a new task across large numbers of potential users and devices.
One drawback of our current method is that there is no scope for added tasks to benefit from each other. Only the features learned for the initial task, such as the ImageNet pre-training, are re-used and adapted for new tasks.
Apart from addressing this issue, another interesting area for future work is the extension to tasks such as object detection that require specialized layers, and expanding existing layers with more capacity as dictated by the task and accuracy targets.

\bibliographystyle{splncs}
\bibliography{egbib}

\begin{thebibliography}{10}

\bibitem{simonyan14VGG}
Simonyan, K., Zisserman, A.:
\newblock Very deep convolutional networks for large-scale image recognition.
\newblock CoRR \textbf{abs/1409.1556} (2014)

\bibitem{ILSVRC15}
Russakovsky, O., Deng, J., Su, H., Krause, J., Satheesh, S., Ma, S., Huang, Z.,
  Karpathy, A., Khosla, A., Bernstein, M., Berg, A.C., Fei-Fei, L.:
\newblock {ImageNet Large Scale Visual Recognition Challenge}.
\newblock IJCV (2015)

\bibitem{french1999catastrophic}
French, R.M.:
\newblock Catastrophic forgetting in connectionist networks.
\newblock Trends in cognitive sciences \textbf{3}(4) (1999)  128--135

\bibitem{kirkpatrick2017overcoming}
Kirkpatrick, J., Pascanu, R., Rabinowitz, N., Veness, J., Desjardins, G., Rusu,
  A.A., Milan, K., Quan, J., Ramalho, T., Grabska-Barwinska, A.,  et~al.:
\newblock Overcoming catastrophic forgetting in neural networks.
\newblock PNAS (2017)

\bibitem{rannen2017encoder}
Rannen, A., Aljundi, R., Blaschko, M.B., Tuytelaars, T.:
\newblock Encoder based lifelong learning.
\newblock In: ICCV. (2017)

\bibitem{li2016learning}
Li, Z., Hoiem, D.:
\newblock Learning without forgetting.
\newblock In: ECCV. (2016)

\bibitem{mallya2017packnet}
Mallya, A., Lazebnik, S.:
\newblock {PackNet}: Adding multiple tasks to a single network by iterative
  pruning.
\newblock arXiv:1711.05769 (2017)

\bibitem{WahCUB_200_2011}
Wah, C., Branson, S., Welinder, P., Perona, P., Belongie, S.:
\newblock {The Caltech-UCSD Birds-200-2011 Dataset}.
\newblock Technical Report CNS-TR-2011-001, California Institute of Technology
  (2011)

\bibitem{krause20133d}
Krause, J., Stark, M., Deng, J., Fei-Fei, L.:
\newblock 3d object representations for fine-grained categorization.
\newblock In: CVPRW. (2013)

\bibitem{Nilsback08}
Nilsback, M.E., Zisserman, A.:
\newblock Automated flower classification over a large number of classes.
\newblock In: ICCVGIP. (2008)

\bibitem{saleh2015large}
Saleh, B., Elgammal, A.:
\newblock Large-scale classification of fine-art paintings: Learning the right
  metric on the right feature.
\newblock In: ICDMW. (2015)

\bibitem{eitz2012humans}
Eitz, M., Hays, J., Alexa, M.:
\newblock How do humans sketch objects?
\newblock In: SIGGRAPH. (2012)

\bibitem{he2016deep}
He, K., Zhang, X., Ren, S., Sun, J.:
\newblock Deep residual learning for image recognition.
\newblock In: CVPR. (2016)

\bibitem{zagoruyko2016wide}
Zagoruyko, S., Komodakis, N.:
\newblock Wide residual networks.
\newblock In: BMVC. (2016)

\bibitem{huang2017densely}
Huang, G., Liu, Z., van~der Maaten, L., Weinberger, K.Q.:
\newblock Densely connected convolutional networks.
\newblock In: CVPR. (2017)

\bibitem{rosenfeld2017incremental}
{Rosenfeld}, A., {Tsotsos}, J.K.:
\newblock Incremental learning through deep adaptation.
\newblock arXiv:1705.04228 (2017)

\bibitem{rebuffi2017learning}
Rebuffi, S.A., Bilen, H., Vedaldi, A.:
\newblock Learning multiple visual domains with residual adapters.
\newblock In: NIPS. (2017)

\bibitem{bilen2016integrated}
Bilen, H., Vedaldi, A.:
\newblock Integrated perception with recurrent multi-task neural networks.
\newblock In: NIPS. (2016)

\bibitem{caruana1998multitask}
Caruana, R.:
\newblock Multitask learning.
\newblock In: Learning to learn.
\newblock (1998)

\bibitem{kokkinos2016ubernet}
Kokkinos, I.:
\newblock Ubernet: Training a universal convolutional neural network for low-,
  mid-, and high-level vision using diverse datasets and limited memory.
\newblock In: CVPR. (2017)

\bibitem{shmelkov2017incremental}
Shmelkov, K., Schmid, C., Alahari, K.:
\newblock Incremental learning of object detectors without catastrophic
  forgetting.
\newblock In: ICCV. (2017)

\bibitem{lee2017overcoming}
Lee, S.W., Kim, J.H., Ha, J.W., Zhang, B.T.:
\newblock Overcoming catastrophic forgetting by incremental moment matching.
\newblock In: NIPS. (2017)

\bibitem{han2015learning}
Han, S., Pool, J., Tran, J., Dally, W.:
\newblock Learning both weights and connections for efficient neural network.
\newblock In: NIPS. (2015)

\bibitem{fernando2017pathnet}
Fernando, C., Banarse, D., Blundell, C., Zwols, Y., Ha, D., Rusu, A.A.,
  Pritzel, A., Wierstra, D.:
\newblock Path{N}et: Evolution channels gradient descent in super neural
  networks.
\newblock arXiv:1701.08734 (2017)

\bibitem{rusu2016progressive}
Rusu, A.A., Rabinowitz, N.C., Desjardins, G., Soyer, H., Kirkpatrick, J.,
  Kavukcuoglu, K., Pascanu, R., Hadsell, R.:
\newblock Progressive neural networks.
\newblock arXiv:1606.04671 (2016)

\bibitem{courbariaux2015binaryconnect}
Courbariaux, M., Bengio, Y., David, J.P.:
\newblock Binaryconnect: Training deep neural networks with binary weights
  during propagations.
\newblock In: NIPS. (2015)

\bibitem{hubara2016binarized}
Hubara, I., Courbariaux, M., Soudry, D., El-Yaniv, R., Bengio, Y.:
\newblock Binarized neural networks.
\newblock In: NIPS. (2016)

\bibitem{li2016ternary}
Li, F., Zhang, B., Liu, B.:
\newblock Ternary weight networks.
\newblock arXiv:1605.04711 (2016)

\bibitem{zhu2016trained}
Zhu, C., Han, S., Mao, H., Dally, W.J.:
\newblock Trained ternary quantization.
\newblock (2017)

\bibitem{guo2016dynamic}
Guo, Y., Yao, A., Chen, Y.:
\newblock Dynamic network surgery for efficient dnns.
\newblock In: NIPS. (2016)

\bibitem{zhou2017places}
Zhou, B., Lapedriza, A., Khosla, A., Oliva, A., Torralba, A.:
\newblock Places: A 10 million image database for scene recognition.
\newblock TPAMI (2017)

\bibitem{long2015fully}
Long, J., Shelhamer, E., Darrell, T.:
\newblock Fully convolutional networks for semantic segmentation.
\newblock In: CVPR. (2015)

\bibitem{SegmentationSplits}
BerekeleyVision:
\newblock Segmentation data splits.
\newblock
  \url{https://github.com/shelhamer/fcn.berkeleyvision.org/tree/master/data/pascal}
  Accessed: 2018-03-11.

\end{thebibliography}
\end{document}